\documentclass[11pt,a4paper]{article}

\usepackage[T1]{fontenc}
\usepackage[utf8]{inputenc}
\usepackage{lmodern}
\usepackage[margin=2.5cm]{geometry}
\usepackage{graphicx}
\usepackage{booktabs}
\usepackage{multirow}
\usepackage{array}
\usepackage{amsmath}
\usepackage{siunitx}
\usepackage{caption}
\usepackage{float}
\usepackage{enumitem}
\usepackage[hidelinks]{hyperref}
\usepackage{url}

\title{\bfseries Detection of Potato Plant Leaf Diseases using
Convolutional Neural Networks}

\author{%
  Haimanot Bitew Alemayehu\textsuperscript{1},
  Simachew Melaku Ayalew\textsuperscript{2},\\[0.2em]
  Chilotie Desalew Mengesha\textsuperscript{3},
  Ayisheshim Almaw Bogale\textsuperscript{4}\\[0.6em],
  Anteneh Birhan Yenew\textsuperscript{5},
  Abrham Tegegne Bogale\textsuperscript{6}\\[0.6em]
  \small\textsuperscript{6,5,4,3,2,1}College of Engineering \& Technology,
  Injibara University, Ethiopia\\
  \small\textsuperscript{1}{haimanot.bitew@inu.edu.et},
  \small\textsuperscript{2}{simachew.melaku@inu.edu.et}\\
  \small\textsuperscript{3}{chilotie.dessalew@inu.edu.et},
  \small\textsuperscript{4}{ayisheshim.almaw@inu.edu.et}\\
  \small\textsuperscript{5}{anteneh.birhan@inu.edu.et},
  \small\textsuperscript{6}{abrham.tegegne@inu.edu.et}\\
 }
\date{\today}
 
\begin{document}
\maketitle

%------------------------- ABSTRACT ---------------------------------
\begin{abstract}
\noindent
Potato is a food‑security crop in Ethiopia but production is affected by foliar diseases, especially early blight and late blight. In farming areas disease identification still depends on visual assessment or laboratory diagnosis approaches that can be slow or difficult to access. This study develops a convolutional neural network (CNN) for recognizing three potato leaf conditions: healthy leaves, early blight and late blight. The proposed model was trained using 2,215 images from PlantVillage. Subsequently assessed with an independent field dataset containing 247 images collected from potato farms in Banja Awi Zone, Ethiopia. On PlantVillage the model obtained 99.04\% training accuracy, 98.21\% validation accuracy and 98.66\% test accuracy. On the field dataset, which represents controlled image‑capture conditions the model achieved 90.28\% accuracy. The study also compares the proposed network with VGG16. Mobilenetv2 using classification performance and computational characteristics. The proposed model contains 446,787 parameters and has an approximate size of 1.70 MB. The model’s training time was shorter, than the training time of VGG16 and MobileNetV2. These results indicate that a compact CNN can provide potato disease classification while reducing computational requirements with potential to benefit smallholder farmers in Ethiopia and similar agricultural contexts.

\end{abstract}


%=====================================================================
\section{Introduction}
%=====================================================================
Agriculture is a part of food production and livelihoods in Ethiopia and potato is one of the important root and tuber crops grown in the country. Potato production however faces constraints, including foliar diseases. Early blight, caused by \textit{Alternaria solani} and late blight, caused by \textit{Phytophthora infestans} are especially important because infections can reduce crop productivity. Previous work in Awi Zone has also reported the presence of late blight in potato‑growing districts \cite{ref1}.

In farming communities people identify potato disease by looking at the leaves. Laboratory tests can give proof but are not always easy to get. Image‑based automated classification can give preliminary identification from leaf images.

Another challenge is that early and late blight symptoms look similar especially as they develop. This similarity makes it hard to tell them apart. Can influence disease management decisions \cite{ref2}.

CNNs are widely used for image‑based plant disease recognition because the convolutional layers can learn patterns directly from training images. Researchers have used them for crops and disease types including potato leaf conditions \cite{ref3}.

For potato disease classification Mahapatra and Panda achieved 96.28\% accuracy with a CNN on PlantVillage data \cite{ref3}. Hoque et al. Achieved 99.14\% accuracy with a lightweight CNN \cite{ref4}. Dey et al. Described a CNN with 204,227 trainable parameters and reported 98.6\% test accuracy \cite{ref5}.

These results show the potential of CNNs Performance on controlled image sets does not always match performance on field images. This difference motivates testing with collected images along with benchmark data. This study focuses on a CNN that classifies three classes healthy, early blight and late blight paying close attention to computational efficiency and testing in local field conditions.

The study notice that many existing potato‑disease studies have limitations. They often use datasets with uniform backgrounds and lighting. These conditions make classification easier than when photos have varying light, background, leaf angle and symptom appearance. Also evidence specific to potato disease in farms is limited.

So the study. Tested a lightweight CNN that recognizes the three chosen potato leaf conditions and looked at how the model behaves on a benchmark dataset and on images taken from farms in Awi Zone.

%=====================================================================

\section{Related Works}

%=====================================================================

The study have seen that recent research focuses more on learning for plant disease recognition. In a review Pacal et al. Covered many studies on deep‑learning methods for plant diseases and highlighted CNNs as a main way to classify images \cite{ref7}. CNNs are useful because they learn feature representations from images during training of needing all visual descriptors to be made manually \cite{ref8}.

\subsection{CNN-Based Approaches for Potato Disease Identification}

The study see that studies on potato leaf classification have used custom CNNs and transfer‑learning models. Mahapatra and Panda got 96.28 achieved accuracy using a CNN on PlantVillage images \cite{ref3}. Hoque et al. Got 99.14 achieved accuracy using a CNN and included regularization techniques, like dropout and batch normalization \cite{ref4}. Dey et al. Created a CNN with 204,227 parameters and said the test accuracy was 98.6 percent \cite{ref5}.

Mulugeta et al. Evaluated a CNN on PlantVillage images and field samples from Ethiopia. Their study reported 85\% accuracy 87\% precision, 85\% recall and 86\% F1‑score on the data \cite{ref6}. These results show the performance gap that can happen when a model shifts, from controlled

benchmark images to field observations.

\subsection{Transfer Learning in Potato Leaf Disease Classification}

Transfer learning is an used strategy for plant disease recognition. Architectures such as VGG16, MobileNetV2, ResNet50 InceptionV3 and DenseNet121 have been employed in image classification studies~\cite{ref9}. Ensemble methods have also merged pretrained networks~\cite{ref10}. Pretrained models can provide feature representations but their parameter counts and computational significantly vary a lot. This difference matters a lot when running them on hardware, with power. Transfer learning stays important because it helps balance how well a model performs with how resources it uses.

\subsection{Real-World Application of CNN Models}

A frequent concern in plant-disease classification is whether high benchmark accuracy can transfer to images taken in farm settings. Studies that use PlantVillage-style images might benefit from controlled capture conditions. For instance KWATCHNet reportedly achieved 99.81\% on PlantVillage data and 93.34\% on field-collected images~\cite{ref12}. This difference shows why an independent field dataset is valuable when testing generalization. Plant-disease classification must therefore be validated in real-world scenarios.

Research on potato disease recognition in Ethiopia is still growing. Existing studies have explored CNN architectures and the classification of blight symptoms~\cite{ref13}. Further investigation that uses collected images and computationally efficient models can help determine how well automated approaches work under regional farming conditions~\cite{ref14}.

%=====================================================================

\section{Methodology}

%=====================================================================

An experimental design was used to develop, train and evaluate the proposed CNN. The experimental procedure involved dataset preparation, image preprocessing and augmentation network construction, model training, performance measurement and comparison with transfer-learning baselines. Both benchmark images and collected field images were included. The methodology covers dataset collection and preprocessing, model architecture design, training and hyperparameter optimization and thorough evaluation using both benchmark and collected datasets.

\subsection{Dataset}

The main training and initial evaluation dataset includes 2{,}215 potato leaf images taken from PlantVillage~\cite{ref15}. These images belong to three classes: early blight and late blight. PlantVillage images were captured under controlled conditions with consistent lighting and uniform background.

\subsection{Field Validation Dataset}

To assess performance under farm conditions 247 extra potato leaf photographs were collected from farms in Banja district Awi Zone, Ethiopia~\cite{ref16}. These photographs were taken with smartphones in light. The images belong to the three target classes and their labels were verified by an agricultural extension expert. The field dataset was kept apart, from model training and validation. Was used for final field testing.

\subsection{Data Preprocessing}
Preprocessing was applied to make the images consistent with the CNN
input requirements and to support stable model training. The same
general preprocessing procedure was used for the benchmark and field
images.
\begin{itemize}[leftmargin=1.4em,itemsep=2pt,topsep=3pt]
  \item \textbf{Image resizing:} each photograph was converted to the
        fixed spatial dimensions required by the network.
  \item \textbf{Pixel normalization:} image intensity values were
        transformed to the interval $[0,1]$ before being supplied to
        the model.
  \item \textbf{Data augmentation:} training images were varied using
        operations such as rotation, horizontal reflection, zooming,
        and brightness modification~\cite{ref17}. These transformations
        were intended to expose the model to greater visual variation
        during learning.
\end{itemize}

\subsection{Proposed CNN Architecture}
The proposed CNN was designed with a focus on a balance between
classification performance and computational cost. Its relatively small
parameter count was intended to make subsequent edge-oriented
deployment more practical.
\begin{figure}[htbp]
\centering
\includegraphics[width=\linewidth]{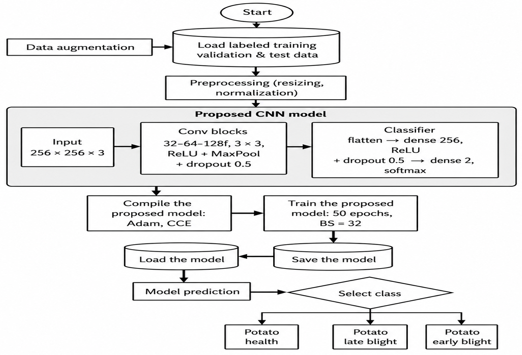}
\caption{Proposed CNN Architecture}
\label{fig:architecture}
\end{figure}

The architecture contains approximately 446{,}787 trainable parameters
and occupies about 1.70~MB.These characteristics make the proposed
network substantially smaller than the compared transfer-learning
configurations~\cite{ref18}, while maintaining a lightweight
architecture.

%=====================================================================
\section{Results}
%=====================================================================
\subsection{Model Training Performance}
During training on PlantVillage, the accuracy curves showed convergence
of the model on the training and validation data. Training accuracy
reached 99.04\%, while validation accuracy reached 98.21\%. The
reported validation loss at the end of training was 0.0833.
\begin{figure}[htbp]
\centering
\includegraphics[width=0.8\linewidth, height=0.4\textheight, keepaspectratio]{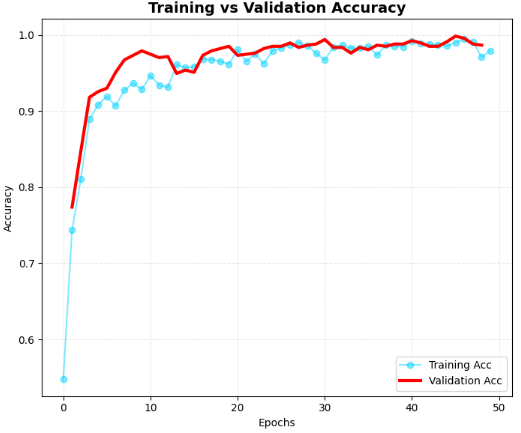}
\caption{Training and Validation Accuracy}
\label{fig:accuracy}
\end{figure}
\begin{figure}[htbp]
\centering
\includegraphics[width=0.8\linewidth, height=0.4\textheight, keepaspectratio]{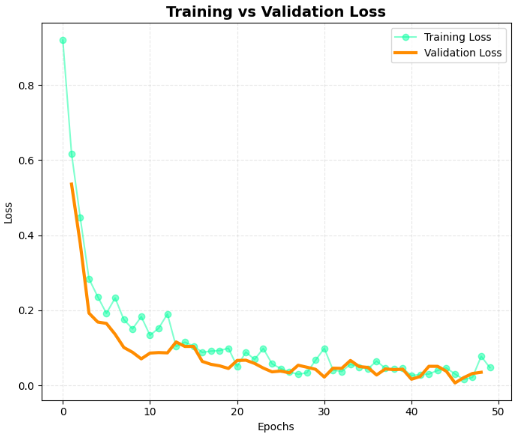}
\caption{Training and Validation Accuracy}
\label{fig:accuracy}
\end{figure}
\subsection{Model Test Performance on PlantVillage}
Evaluation on the held-out PlantVillage test set produced an overall
accuracy of 98.66\%. The confusion matrix (Table~\ref{tab:cm}) indicates
that most samples were assigned to their correct classes, with a small
number of errors distributed among the three categories.

\begin{table}[H]
  \centering
  \caption{Confusion matrix of the proposed model on the PlantVillage
  test set.}
  \label{tab:cm}
  \begin{tabular}{lccc}
    \toprule
    & \textbf{Healthy} & \textbf{Early blight} & \textbf{Late blight}\\
    \midrule
    Healthy       & 215 & 1 & 2 \\
    Early blight  & 1   & 214 & 3 \\
    Late blight   & 2   & 0   & 214 \\
    \bottomrule
  \end{tabular}
\end{table}

\begin{figure}[htbp]
\centering
\includegraphics[width=\linewidth]{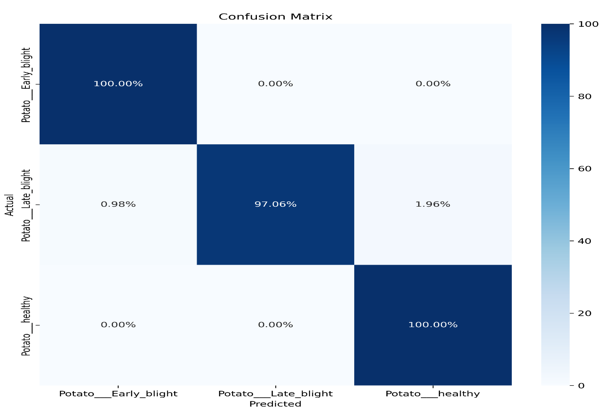}
\caption{Confusion Matrix Visualization}
\label{fig:confusion}
\end{figure}

\subsection{Validation on the Field-Collected Dataset}
The independent Banja field dataset was more varied than the benchmark images. This is because the photographs were taken under light and showed different backgrounds and different ways symptoms appeared. The model achieved 90.28\% accuracy and a loss of 0.4615 on these 247 images
(Table~\ref{tab:field}).

\begin{table}[H]
  \centering
  \caption{Performance on the field-collected dataset.}
  \label{tab:field}
  \begin{tabular}{ll}
    \toprule
    \textbf{Metric} & \textbf{Value} \\
    \midrule
    Accuracy        & 90.28\% \\
    Loss            & 0.4615 \\
    Dataset size    & 247 images \\
    \bottomrule
  \end{tabular}
\end{table}

\subsection{Comparison, with Transfer Learning Models}
In this section we compare the proposed CNN with transfer learning models. The proposed CNN was evaluated alongside VGG16 and MobileNetV2 after the comparison models had been fine‑tuned using the potato‑disease classification task. The results are presented in Tables~\ref{tab:params} and~\ref{tab:perf} allowing accuracy, loss, parameter count, model size and training time to be examined together.

\begin{table}[H]
  \centering
  \caption{Comparative model parameters and efficiency.}
  \label{tab:params}
  \small
  \begin{tabular}{lcccc}
    \toprule
    \textbf{Model} & \textbf{Total} & \textbf{Trainable} &
    \textbf{Model size} & \textbf{Avg.\ time}\\
    & \textbf{params} & \textbf{params} & \textbf{(MB)} &
    \textbf{/epoch (s)}\\
    \midrule
    VGG16         & 2{,}261{,}827 & 300{,}843 & 8.63 (total), 0.015 (trainable) & 9.96 \\
    MobileNetV2   & 14{,}716{,}227 & 100{,}539 & 56.14 (total), 1.26 (trainable) & 7.68 \\
    Proposed      & 446{,}787 & 446{,}787 & 1.70 & 5.58 \\
    \bottomrule
  \end{tabular}
\end{table}

For the PlantVillage comparison the model achieved 99.04\%  training accuracy and a validation loss of 0.0268 in the experiment. The losses reported for MobileNetV2 and VGG16 were 0.0916 and 0.1804. These values show differences, in the learning behaviour of the three configurations.

\begin{table}[H]
  \centering
  \caption{Comparative performance on both datasets.}
  \label{tab:perf}
  \small
  \begin{tabular}{llccc}
    \toprule
    \textbf{Dataset} & \textbf{Metric} & \textbf{VGG16} &
    \textbf{MobileNetV2} & \textbf{Proposed}\\
    \midrule
    \multirow{5}{*}{PlantVillage}
      & Precision & 96     & 98     & 96 \\
      & Recall    & 98     & 98     & 99 \\
      & Accuracy  & 97     & 97     & 99.04 \\
      & F1-score  & 97     & 98     & 97 \\
      & Loss      & 0.1804 & 0.0916 & 0.0268 \\
    \midrule
    \multirow{5}{*}{Field dataset}
      & Precision & 84     & 89     & 91 \\
      & Recall    & 83     & 86     & 91 \\
      & Accuracy  & 82.59  & 86.23  & 90.28 \\
      & F1-score  & 83     & 86     & 90 \\
      & Loss      & 0.5691 & 0.3816 & 0.4615 \\
    \bottomrule
  \end{tabular}
\end{table}

The accuracy difference between the PlantVillage test evaluation and the field evaluation is 8.38 percentage points for the proposed model. This accuracy difference shows that the field evaluation images are harder to classify than the controlled benchmark images and gives a measure of the proposed models performance, across different settings.

\subsection{Training Efficiency}

The proposed model took 5.58~seconds per epoch in the comparison. VGG16 took 9.96~seconds. Mobilenetv2 took 7.68~seconds. Therefore within the setup the proposed network needed less training time than both comparison configurations.

%=====================================================================

\section{Discussion}

%=====================================================================

The paper found that the proposed lightweight CNN successfully completed the goal of detecting potato leaf disease. The model performed well on both the PlantVillage dataset and on the field collected dataset.

In the reported experiments the proposed network reached 99.04\% training accuracy, 98.21\% validation accuracy and 98.66\% PlantVillage test accuracy. The field evaluation yielded 90.28\% accuracy. The comparison tables show that the proposed configuration had a validation loss than VGG16 and MobileNetV2 in the same experiment.

The gap between training and validation accuracy was small during the reported training run. Yet the performance on the independent field dataset was lower than on PlantVillage. This difference matches the visual variability that comes from field image capture and it shows that benchmark performance should not be taken as the same as field performance.

The paper observed that the proposed architecture has 446{,}787 parameters and a size of about 1.70~MB. Its epoch time was 5.58~seconds while VGG16 took 9.96~seconds. Mobilenetv2 took 7.68~seconds. These measurements confirm that the proposed architecture is computationally compact in the reported environment.

The paper noted that a compact model can be helpful where computing resources or connectivity are limited. In the field experiment the proposed model achieved 90.28\% accuracy. Mulugeta et al. Reported accuracy, on their collected Ethiopian dataset. Because these accuracy results come from datasets and experimental settings results should be seen as findings, not a direct comparison.
%=====================================================================
\section{Conclusion}
%=====================================================================
This study developed and evaluated a lightweight CNN for classification
of potato leaves into healthy, early blight, and late blight categories.
The proposed model achieved 98.66\% test accuracy on the selected
PlantVillage test set and 90.28\% accuracy on the independently collected
Banja field dataset. The comparison experiments also examined VGG16 and
MobileNetV2 in terms of performance and computational characteristics.

The main contributions of the present study are as follows:
\begin{enumerate}[leftmargin=1.6em,itemsep=3pt,topsep=3pt]
  \item Proposing an automated system for the detection of potato leaf
        diseases using a lightweight CNN.

  \item The proposed configuration recorded 98.66\% PlantVillage test
        accuracy, while the comparative experiment reports a 0.0268
        validation loss for the proposed model. These results describe
        the performance of the proposed configuration under the stated
        experimental setup.

  \item On the Banja field dataset, the proposed model obtained 90.28\%
        accuracy. The difference between the PlantVillage test accuracy
        and the field accuracy was 8.38 percentage points when those two
        test results are compared directly.

  \item The model contains 446{,}787 trainable parameters and has an
        approximate size of 1.70~MB. The reported training-time
        measurements also indicate a lower average epoch time than the
        VGG16 and MobileNetV2 configurations used for comparison.
\end{enumerate}

The findings indicate that lightweight image-classification models may
have practical relevance to precision-agriculture applications. A
field-deployed system could provide preliminary disease classification
from leaf photographs, although operational treatment decisions should
remain supported by appropriate agricultural expertise.

The results indicate potential for further investigation of
deep-learning-based potato disease management in Awi Zone. Future work
will focus on:
\begin{itemize}[leftmargin=1.4em,itemsep=2pt,topsep=3pt]
  \item Increasing the number of Banja field images used for future
        fine-tuning and evaluation.
  \item Extending the class set to include additional potato diseases,
        including bacterial wilt and virus-related symptoms, as well as
        healthy leaves affected by non-disease stress.
  \item Developing a mobile application with a simple user interface and
        local-language support, including Amharic and Awign.
  \item Combining image-based diagnosis with weather information to
        investigate an early-warning framework for disease risk.
  \item Assessing the practical and economic effects of the system
        through field-based evaluation with farmers.
\end{itemize}

%=====================================================================
%  REFERENCES
%  NOTE: several entries in the source manuscript had no author names.
%  Complete these before submission (marked with %% TODO).
%=====================================================================

\end{document}